\documentclass{article} 
\usepackage{collas2022_conference,times}

\usepackage{wrapfig}
\usepackage{graphicx}
\usepackage{subfigure}
\usepackage{booktabs}
\usepackage{caption}
\usepackage{stackengine}
\usepackage{multicol}
\usepackage{multirow}
\usepackage{amssymb}
\usepackage{algorithm}
\usepackage{algpseudocode}

\usepackage{xcolor,comment}
\usepackage{pifont}
\newcommand{\cmark}{\ding{51}}%
\newcommand{\xmark}{\ding{55}}%
 
\usepackage{amsmath,amsfonts,bm}

\def\eqref#1{equation~\ref{#1}}
\def\Eqref#1{Equation~\ref{#1}}

\def\1{\bm{1}}

\DeclareMathAlphabet{\mathsfit}{\encodingdefault}{\sfdefault}{m}{sl}
\SetMathAlphabet{\mathsfit}{bold}{\encodingdefault}{\sfdefault}{bx}{n}

\newcommand{\R}{\mathbb{R}}

\usepackage{hyperref}
\hypersetup{
    colorlinks=true,
    linkcolor=red,
    filecolor=magenta,      
    urlcolor=blue,
    citecolor=purple,
    pdftitle={Overleaf Example},
    pdfpagemode=FullScreen,
    }

\makeatletter
\newcommand{\printfnsymbol}[1]{%
  \textsuperscript{\@fnsymbol{#1}}%
}
\makeatother

\title{Differencing based Self-supervised pretraining for Scene Change Detection }

\author{%
  Vijaya Raghavan T. Ramkumar, Elahe Arani\thanks{Equal advising.~~~~~~~~{$^1$ We open source our code at \url{https://github.com/NeurAI-Lab/DSP}.}}, Bahram Zonooz\printfnsymbol{1} \\
  Advanced Research Lab, NavInfo Europe, The Netherlands \\
  \texttt{\{vijaya.ramkumar, elahe.arani\}@navinfo.eu, bahram.zonooz@gmail.com} \\
}

\collasfinalcopy 

\begin{document}

\maketitle

\begin{abstract}
Scene change detection (SCD), a crucial perception task, identifies changes by comparing scenes captured at different times. SCD is challenging due to noisy changes in illumination, seasonal variations, and perspective differences across a pair of views. Deep neural networks based solutions require a large quantity of annotated data which is tedious and expensive to obtain. On the other hand, transfer learning from large datasets induces domain shift. To address these challenges, we propose a novel  \textit{Differencing self-supervised pretraining (DSP)} method that uses feature differencing to learn discriminatory representations corresponding to the changed regions while simultaneously tackling the noisy changes by enforcing temporal invariance across views. Our experimental results on SCD datasets demonstrate the effectiveness of our method, specifically to differences in camera viewpoints and lighting conditions. Compared against the self-supervised Barlow Twins and the standard ImageNet pretraining that uses more than a million additional labeled images, DSP can surpass it without using any additional data. Our results also demonstrate the robustness of DSP to natural corruptions, distribution shift, and learning under limited labeled data.\footnotemark
\end{abstract}


\section{Introduction}

Scene change detection (SCD) is a critical perception task that helps to identify changes in a scene captured at different times. In recent years, SCD has been gaining popularity in the field of computer vision, robotics, and remote sensing \citep{hamaguchi2019rare, sakurada2020weakly, alcantarilla2018street} as its various real world applications such as ecosystem monitoring, urban expansion, remote surveillance, autonomous driving, and damage assessment have an immensely positive impact on society. For instance, in autonomous driving and robotics applications, the problem of generating and maintaining maps of ever-changing environments is of utmost importance for dynamic localization, and robust operation of vehicles/robots in urban landscapes. SCD helps to alleviate the problem of mapping and efficient maintenance by continuously monitoring the changes of the scene at different time instances \citep{alcantarilla2018street}. Thus, it plays an important role in many real world applications by perceiving the changes occurring in the environment. 

Generally, in SCD, the changed region is smaller than the unchanged region with uncertainty in change location and direction. Moreover, the changed region that needs to be detected depends on the nature of the application and is classified into semantic changes (relevant) and noisy changes (irrelevant). The structural changes caused by the appearance or disappearance of objects present in a scene are considered as semantic changes, while the changes induced by the radiometric (illumination, shadows, seasonal changes) and geometric variations (viewpoint differences caused by camera rotation) are considered as noisy changes \citep{alcantarilla2018street, sakurada2015change, guo2018learning}. A critical challenge in SCD is that these noisy changes are entangled with the semantic changes that alter the appearance of an image, thus degrading the change detection performance \citep{guo2018learning}. 

\begin{wrapfigure}[20]{R}{0.5\textwidth}
     \centering
     \vspace{-18pt}
     \includegraphics[width = 0.53\textwidth, trim={0.7cm 0.4cm 0.7cm 0.7cm},clip]{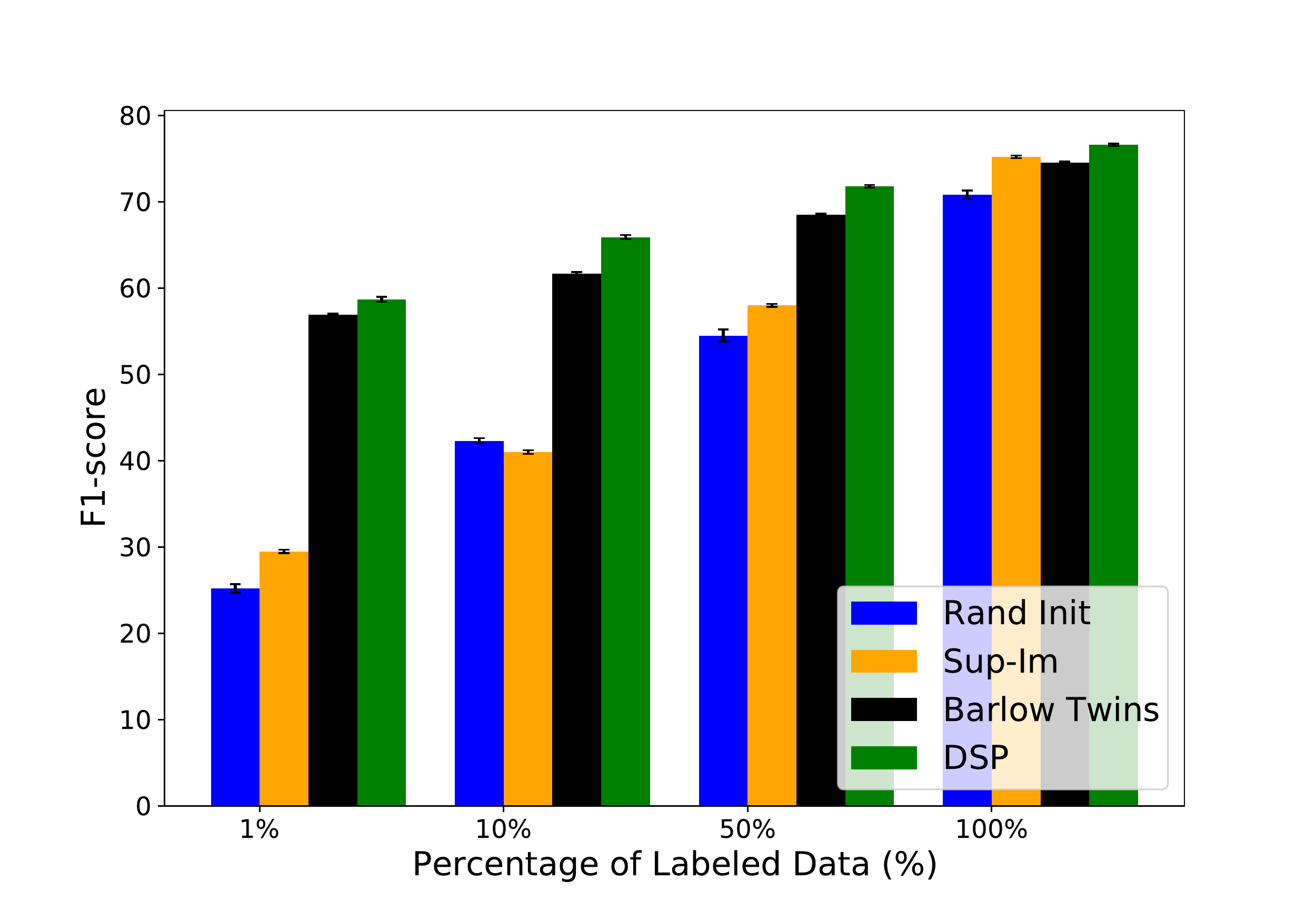}
     \caption{Performance comparisons of supervised and the proposed self-supervised pretraining (DSP) on VL-CMU-CD dataset under limited label scenario.}
      \label{fig:limited labels}
\end{wrapfigure} 

Previous studies based on deep neural networks have proposed to extract multi-level feature representations from the input images to improve the performance of SCD against noisy changes \citep{guo2018learning, alcantarilla2018street, varghese2018changenet, lei2020hierarchical}. However, the success of these state-of-the-art methods hinges on a large quantity of annotated data. For instance, on average, it takes around 20 minutes and 156 minutes to annotate a single pair of images in the panoramic change detection (PCD) \citep{sakurada2013detecting} and panoramic semantic change detection (PSCD) \citep{sakurada2020weakly} dataset, respectively. Therefore, large-scale labeled datasets for SCD are still scarce, and expensive to obtain \citep{shi2020change}. To address the dependency on labeled data, various SCD approaches initially pretrain their models on large-scale datasets such as ImageNet \citep{deng2009imagenet} in a supervised manner and then finetune with a small quantity of pixel-level annotations on domain-specific dataset \citep{guo2018learning, sakurada2020weakly, chen2021dr}. However, there still exists the problem of (1) domain shift as the distribution of the ImageNet data widely differs from that of SCD datasets, (2) nature of feature representation learned by transfer learning models from classification tasks is sub-optimal for SCD. These problems lead to the degradation of change detection performance in SCD methods.

To attenuate the reliance of SCD models on a large amount of dense pixel-level annotations and transfer learning from large-scale labeled out-of-distribution data, we propose a novel self-supervised pretraining approach that utilizes unlabeled data to learn task-specific representations for the downstream task of SCD. Our method, DSP, uses feature differencing to learn discriminatory representations corresponding to the changed regions that are beneficial for the downstream task of SCD. Furthermore, we propose \textit{invariant prediction (IP)} objective and \textit{change consistency regularization (CR)}, together referred to as \textit{temporal consistency (TC)} loss, to reduce the effects of differences in the lighting conditions or camera viewpoints by enhancing the image alignment between the temporal images in the decision and feature space, respectively. With extensive experiments, we show that our proposed approach achieves remarkable performance compared to ImageNet pretraining under limited labels scenario as seen in Figure~\ref{fig:limited labels}. To the best of our knowledge, this is the first work on SCD that relaxes the requirement of large-scale annotated datasets and the need to pretrain on additional large-scale labeled data in a computationally efficient way. Our contribution can be summarized as follows:
\begin{itemize}
\item We propose a \textit{differencing based self-supervised pretraining (DSP)} method that learns change representations (task-specific) relevant for scene change detection.
\item We propose an \textit{invariant prediction (IP)} objective and change \textit{consistency regularization (CR)} to mitigate the effect of noisy changes across a pair of views.
\item We evaluated the proposed methods on two challenging SCD datasets. DSP surpasses the widely used ImageNet pretraining without any additional data. Also, DSP pretraining enhances the SCD performance compared to the standard Barlow Twins \citep{zbontar2021barlow} method.
\item Current scene change detection models are vulnerable to severe performance impairments on images with natural corruptions, and the proposed self-supervised pretraining significantly enhances the robustness of the model to natural corruptions.
\item The effectiveness of the proposed self-supervised pretraining under limited labels and generalization to out-of-distribution data is verified.
\end{itemize}

\section{Related Work}

\textbf{Scene Change Detection (SCD)} 

Recently, deep neural networks have demonstrated remarkable performance in SCD tasks when compared to the traditional change detection methods \citep{guo2018learning, alcantarilla2018street, varghese2018changenet, ramkumar2021self, sakurada2020weakly}. \citet{alcantarilla2018street} propose a change detection method called CDNet that utilizes CNNs to extract dense geometry and accurate registration to warp images from different times for the change detection. \citet{guo2018learning} proposes a supervised network using contrastive loss to learn the discriminative features with the customized feature distance metrics. Additionally, they also propose a threshold contrastive loss function to tackle significant viewpoint differences present in the input image pairs. On the other hand, \citet{sakurada2017dense} and \citet{bu2020mask} integrate dense optical flow methods with CNNs to model the spatial correspondences between images that minimize the noise due to significant viewpoint differences. Furthermore, \citet{sakurada2020weakly} also proposes a method to capture the multi-scale feature information using hierarchically dense connections for semantic change detection. DR-TANet \citep{chen2021dr} proposes a lightweight network that utilizes a temporal self-attention mechanism to enhance the feature correlation between the temporal images.

Despite the advances in SCD, all methods hinge heavily on the availability of large-scale manually annotated datasets that are hard to obtain. When the labeled data is limited, they depend on transfer learning from models pretrained in a supervised manner on some other big datasets such as ImageNet \citep{sakurada2020weakly}. Yet, transfer learning from the ImageNet pretraining models induces domain shift that reduces the change detection performance in the downstream task. This work proposes a self-supervised pretraining method that utilizes in-distribution unlabeled data to learn representations relevant to SCD. Moreover, our proposed approach is simple and can be easily adapted to any application of change detection.

\textbf{Self-supervised Representation Learning (SSL)}

Contrastive learning has recently gained popularity because of its ability to learn useful representations from unlabeled data. In contrastive learning, the output embedding of the sample along with its augmented version (positive pair) are pulled close to each other while contrastive samples (negative pairs) are pushed away \citep{chen2020simple, tian2020contrastive, chen2020big, he2020momentum}. Many SSL methods employ InfoNCE loss \citep{wu2018unsupervised, bhat2021distill} to achieve contrastive learning during model training and yield good performance. However, these methods require a large number of negative samples to prevent the model from learning trivial constant solutions. To sample large number of contrastive pairs, MoCo \citep{he2020momentum} uses memory banks whereas SimCLR \citep{chen2020simple} uses bigger batch size to sample contrastive pairs. However, this imposes an additional memory constraint during self-supervised training. To address this issue of trivial solutions, several techniques have been proposed in the literature. BYOL \citep{grill2020bootstrap}, SimSiam \citep{chen2021exploring} avoid trivial solutions by introducing asymmetric network architecture using 'predictor' network and asymmetric parameter updates using momentum encoder and stop-gradient. Unlike BYOL and SimSiam, where asymmetric network or parameter updates are required to avoid trivial solutions, Barlow Twins \citep{zbontar2021barlow}, on the other hand, prevent collapse by maximizing the information content of the embeddings. They use cross-correlation to maximize the correlation between the distorted views of samples to be close to the identity while minimizing the redundancy between the components of these vectors. Therefore, we employ Barlow Twins as a SSL objective in our framework since it does not require large negative samples or asymmetric network or stop gradient to avoid trivial solutions.    

To date, most existing self-supervised pretraining methods are designed to learn representations by bringing the positive pairs closer \citep{grill2020bootstrap, chen2020simple, zbontar2021barlow, chen2021exploring}. These pretrained models may be sub-optimal for dense change detection task as it requires high visual correspondence of unchanged regions and low correspondence of changed region between image pairs taken at different times. Eventually, considering the temporal images as a positive pair and pulling them closer may diminish the discrimination ability of the model in the downstream SCD task, because the representations of the two images along with that of the changed regions are forced to be closer together. 
Hence, inspired by the recent improvements in self-supervised representation learning, we propose a novel feature differencing-based self-supervised pretraining approach that learns better representations of the changed regions directly from unlabeled data. Thus, we make self-supervised framework more suitable for SCD.

\section{Method}\label{methodology}

SCD aims to distinguish the changed and unchanged pixels of image pairs captured at different times. Let us consider two images $T_0$ and $T_1$ acquired over the same geographical region at two different times. Our goal is to generate a change intensity map that contains the most salient changed pixels from multi-view images $T_0$ and $T_1$ in a label-efficient way. To achieve that, we first pretrain a model on a set of unlabeled images $\{I_{UL} = (T_0, T_1)_i\}_{i=1}^N$ in a self-supervised manner and then finetune it on a small set of labeled image pairs $\{I_L = (T_0,T_1, L)_i\}_{i=1}^M$. We propose a feature differencing-based self-supervised approach that learns features which are invariant to noisy changes.     
Our proposed pretraining method and objective functions are described in detail in Sections~\ref{sec:dsscd} and \ref{sec:loss_fn}, and the SCD algorithm used to evaluate the proposed pretraining methods is presented in Section~\ref{changedetection_setup}.

\begin{figure}[!tb]
\centering
  \includegraphics[width=.87\textwidth, trim={0cm 0cm 1.8cm 0cm},clip]{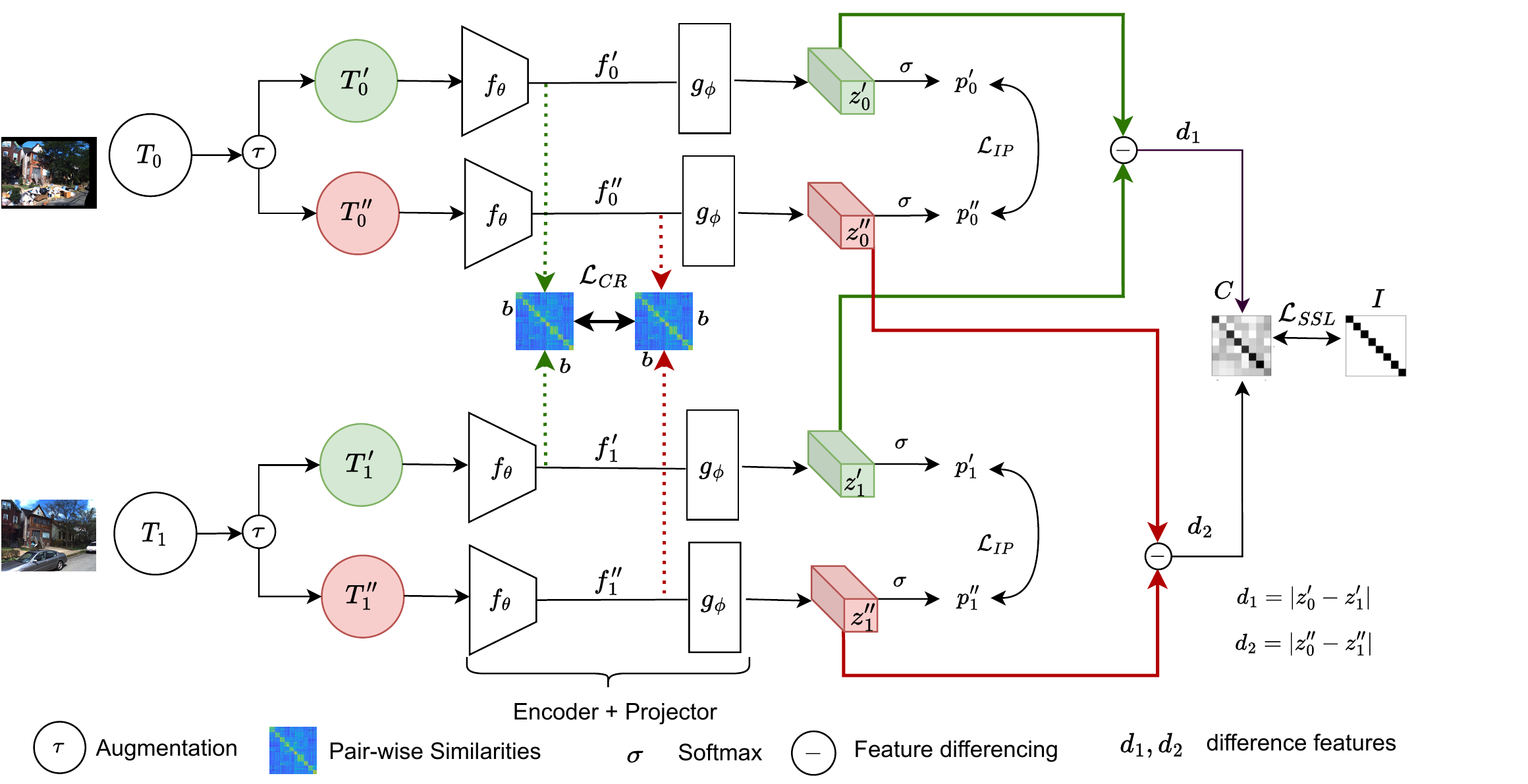}
\caption{Schematics of proposed \textit{differencing based self-supervised pretraining (DSP)} method for scene change detection. DSP uses absolute feature differencing to learn the representation of the changed region directly from the unlabelled images while simultaneously incorporating \textit{temporal consistency} between image pairs taken at different times through an invariant prediction objective $\mathcal{L}_{IP}$  and change consistency regularization $\mathcal{L}_{CR}$.}
\label{fig:method}
\end{figure}

\subsection{Differencing based Self-supervised Pretraining for Scene Change Detection (DSP)}\label{sec:dsscd}

 We propose a novel differencing based Self-supervised pretraining for SCD called DSP that maximizes the correlation of the difference (changed) regions across temporal views to learn distinctive representation. DSP gets an image pairs ($T_0$, $T_1$) from different time instances as input. Then, random photometric transformations are applied to each input images to obtain two pairs of augmented images $T_0 \to (T_0',T_0'')$ and $T_1 \to (T_1',T_1'')$. These augmented images are passed into the same encoder ($f_{\theta}$) and projection head ($g_{\phi}$) to output corresponding feature representations. To learn the representation of the changed features between the pair of images directly, absolute feature differencing is applied over the outputs of the projection head to obtain difference representations ($d_1$, $d_2$):
\begin{align}
\label{eqn:differencing}
\begin{split}
 d_1 = \lvert g(f({\mathbf{T_0'}})) - g(f({\mathbf{T_1'}}))\rvert, ~~~~~
 d_2 = \lvert g(f({\mathbf{T_0''}})) - g(f({\mathbf{T_1''}}))\rvert
\end{split}
\end{align}
We constrain our model under the assumption that semantic changes between the image features ($T_0',T_1'$) and ($T_0'',T_1''$) should remain the same irrespective of the applied augmentations., ie $d_1 \simeq d_2$. Therefore, a Self-supervised objective function ($\mathcal{L}_{SSL}$) (See Section~\ref{sec:loss_fn}) is applied on the difference representations $d_1$ and $d_2$ to maximize the cross-correlation of the changed features. In this way, the model is encouraged to learn task-specific information about the relevant changes that occur between the image pairs. After the pretraining step, the parameters of the encoder $f_\theta$ are transferred to the downstream task of change detection.

\subsection{Loss Function} \label{sec:loss_fn}
This section describes the three objective functions that the proposed DSP uses for training: Self-supervised loss ($\mathcal{L}_{SSL}$), invariant prediction loss ($\mathcal{L}_{IP}$), change consistency regularization ($\mathcal{L}_{CR}$).

\textbf{Self-supervised Loss.}
The DSP network is trained in a self-supervised manner using the objective function proposed by \citet{zbontar2021barlow}. Unlike \citet{zbontar2021barlow} where they maximize the cross-correlation of the transformed views of the same image to be closer to the identity matrix, we maximize the information of the difference representations ($d_1, d_2$) between the corresponding image pair taken at different times to be similar in the feature space:
\begin{equation}
  \label{btloss}
   \mathcal{L}_{SSL} \triangleq  \underbrace{\sum_i (1-C_{ii}^2)}_\text{Invariance term} +\underbrace{\lambda \sum _i \sum_{j\neq i}C_{ij}^2}_\text{Redundancy reduction term}
\end{equation}
\begin{equation}
    \label{cross-corr}
    {C}_{ij} \triangleq \frac{\sum_b (d_1)_{b,i}(d_2)_{b,j} }{\sqrt{\sum _b ((d_1)_{b,i})^2}\sqrt{\sum _b ((d_2)_{b,j})^2}}
\end{equation}
where $\lambda$ is a trade-off constant between invariance and redundancy reduction term, $C$ is the cross-correlation matrix calculated between the difference representations $d_1$ and $d_2$ with $b$ samples of the batch, and $i$, $j$ index the vector dimension of the network outputs. This objective function consists of two components: (1) the invariance term that makes the difference representations of the augmented input image pair $(T_0', T_1')$, $(T_0'', T_1'')$ invariant to the presence of noisy changes (e.g., illuminations) by maximizing the diagonal components of the cross-correlation matrix close to identity matrix. (2) the redundancy reduction term tries to decorrelate the off-diagonal components of the cross-correlation matrix and thus, aligning the difference representations ($d_1,d_2$) to be similar.

\textbf{Invariant prediction.}
Inspired by the observation that the predictions of semantically similar samples should be similar irrespective of different lighting conditions. To achieve invariant prediction, we propose to explicitly enforce invariance under augmentations between intra-view ($T_0',T_0''$) and ($T_1',T_1''$) features. To this end, we employ the Jensen-Shannon divergence $\mathcal{D}_{JS}$ as an objective to achieve invariant prediction:

\begin{align}
\label{eqn:intra_TA_loss}
\begin{split}
 \mathcal{D}_{JS}^{T_0} &= 1/2 *( \mathcal{D}_{KL}(\sigma(z_0')||M_1) + \mathcal{D}_{KL}(\sigma(z_0'')||M_1) )\\
 \mathcal{D}_{JS}^{T_1} &= 1/2 *( \mathcal{D}_{KL}(\sigma(z_1')||M_2) + \mathcal{D}_{KL}(\sigma(z_1'')||M_2) )\\
 \mathcal{L}_{IP} &=  \mathcal{D}_{JS}^{T_0} +  \mathcal{D}_{JS}^{T_1}
 \end{split}
\end{align}

Where $M_1 = 1/2 * (\sigma(z_0') + \sigma(z_0''))$; $M_2 = 1/2 * (\sigma(z_1') + \sigma(z_1''))$ are the mean probabilities belonging to the corresponding features. $\sigma$ is the softmax function, and $\mathcal{D}_{KL}$ is the Kullback–Leibler divergence. Using $\mathcal{L}_{IP}$, we force the DSP model to learn features invariant to lighting conditions.

\textbf{Change consistency regularization.}
Generally, the task of SCD involves image pairs captured at different times. These image pairs are predominantly affected by the noisy changes in illumination, seasonal variations, and view-point differences which makes it challenging to detect the relevant changes. Therefore, it is important to enforce similarity in the feature space of temporal image pairs as the prior probability of occurrence of change is less compared to no change between the images. To achieve this, we propose a change consistency regularization that preserves the semantic similarity between temporal images in the feature space. Similar to \citet{tung2019similarity} where they preserve similarity of activations between the teacher and student network as a way of knowledge distillation, we enforce the encoder of the DSP network to elicit similar activations for the image pairs taken at different times. In this way, we incorporate temporal invariance into the DSP model which implicitly makes the model robust to noisy changes during the pretraining stage.

Given an input mini-batch, the activation map produced by the encoder of the DSP network for transformed $T_0$ images is given by $f_0' \in \R^{b\times c \times h \times w}$ and $f_0'' \in \R^{b\times c \times h \times w}$, where $b$ is the batch size, $c$ is the number of output channels and $h$ and $w$ are the spatial dimensions. Similarly, the activation map produced by the encoder for transformed $T_1$ images are denoted as $f_1'$ and $f_1''$. Note that the dimensions of the activation maps for both the image pairs will be same as they are produced by the same encoder. To incorporate the temporal consistency across $T_0$ and $T_1$ images in the feature space, we construct a pair-wise similarity between $f_0'$ and $f_1'$ given by, $G' = <f_0', f_1'>$. Analogously, let $G'' = <f_0'', f_1''>$ be the pair-wise similarity matrix between $f_0''$ and $f_1''$. Then, we define the change consistency regularization as:
\begin{align}
\label{eqn:changeloss}
\begin{split}
\mathcal{L}_{CR} = \frac{1}{b^2} ||G' - G''||_F^2
\end{split}
\end{align}
Where $||.||_F $ is the Frobenius norm and $<.,.>$ is the dot product.
 
\textbf{Overall loss}: We combine the proposed invariant prediction loss $\mathcal{L}_{IP}$ and change consistency regularization $\mathcal{L}_{CR}$ together and refer to as temporal consistency objective $\mathcal{L}_{TC}$. Therefore, the overall objective for the DSP network obtained as the weighted sum of self-supervised and temporal consistency objectives:
\begin{align}
\label{total_loss}
\begin{split}
\mathcal{L}_{TC} &= \mathcal{\alpha} \mathcal{L}_{IP} + \mathcal{\beta}\mathcal{L}_{CR}\\
\mathcal{L}_{Total} &= \mathcal{L}_{SSL} + \mathcal{L}_{TC} 
\end{split}
\end{align}
where $\alpha$ and $\beta$ are the loss balancing weights.

\section{Experiments}
\subsection{SCD Datasets}
To train and validate the proposed framework, we considered two SCD datasets subjected to noisy changes such as illumination, shadows, seasonal variations, and camera viewpoint differences. 

\textbf{VL-CMU-CD dataset} \citep{alcantarilla2018street}: It consists of 152 perspective image sequences taken at different time instances. Each image sequence contains approximately nine pairs of softly co-registered images taken at different times. Therefore, it has 1362 image pairs of 1024$\times$768 are generated with their manually labeled pixel-level change masks. This dataset portrays the typical macroscopic changes that occur in an urban scenario. These changes are hard to distinguish as this dataset contains low-resolution images that are predominantly affected by noisy changes.

\textbf{Panoramic Change Detection (PCD) dataset} \citep{sakurada2013detecting}: It contains two subsets of data, namely 'TSUNAMI' and 'GSV'. Each subset has 100 pairs of non-registered panoramic images ($224 \times 1024$ pixels) along with the manually labeled change masks. TSUNAMI subset contains image pairs representing the aftermath of tsunami-affected areas in Japan, whereas the GSV subset contains image pairs belonging to Google street view. Compared to the VL-CMU-CD dataset, this dataset contains good resolution images and is affected by noisy changes, and contains large viewpoint differences. 

In both the datasets, the structural changes such as the emergence/vanishing of buildings and cars are considered relevant, and the noisy changes are deemed irrelevant and excluded from the ground truth change map.

\subsection{Evaluation Criteria}
We use the F1-score metric to evaluate the change detection performance after finetuning. The value of the F1-score ranges from 0 to 1. The higher the F1-score, the better the precision and recall. 
\begin{equation}
    \label{eq3}
    F1-score = \frac{2\cdot Recall \cdot Precision}{Recall + Precision}
\end{equation} 

\section{Results and Discussion}
Experiments are conducted by finetuning the SCD model on DR-TANet \citep{chen2021dr} with four sets of pretraining strategies on PCD and VL-CMU-CD datasets. (1) Random initialized (Rand Init), (2) Supervised ImageNet pretraining (Sup-Im), (3) Standard Barlow Twins self-supervised pretraining, (4) Randomly initialized proposed self-supervised pretraining (DSP). The details of the DSP pretraining and finetuning set-up is provided in Appendix Section~\ref{setup}

\begin{table}[t]
\caption{Performance of DR-TANet model trained and evaluated on VL-CMU-CD and PCD datasets using different pretraining methods. Precision and recall for Tsunami and GSV are provided in Table \ref{tab:pcd_pr}.}
  \label{res:vlcmucd}
  \centering
\begin{tabular}{lcccccccccccc}
 \toprule
\multirow{3}{*}{Methods} & \multicolumn{3}{c}{\multirow{2}{*}{VL-CMU-CD}} & \multicolumn{5}{c}{PCD dataset} \\  \cmidrule(r){5-9}
 & \multicolumn{3}{c}{} & \multicolumn{1}{c}{Tsunami} & \multicolumn{1}{c}{GSV} & \multicolumn{3}{c}{Average} \\ \cmidrule(r){2-9} 
 & Precision & Recall & F1-score  & F1-score &  F1-score & Precision & Recall & F1-score \\ \midrule
Rand Init  & 73.1\tiny$\pm0.30$ & 68.7\tiny$\pm0.24$  &  70.8\tiny$\pm0.51$  & 63.4\tiny$\pm0.25$  &40.7\tiny$\pm0.21$  &  54.0\tiny$\pm0.27$ &  50.7\tiny$\pm0.26$ &53.5\tiny$\pm0.24$  \\
Sup-Im  &  80.5\tiny$\pm0.31$& 70.6\tiny$\pm0.23$  & 75.2\tiny$\pm0.32$ & 68.7\tiny$\pm0.13$ & 46.5\tiny$\pm0.12$ & 61.6\tiny$\pm0.13$ &  54.6\tiny$\pm0.15$ & 57.6\tiny$\pm0.12$ \\
Barlow Twins & 80.1\tiny $\pm0.12$& 69.5\tiny $\pm0.12$  & 74.5\tiny $\pm0.12$ & 70.9\tiny $\pm0.18$    & 45.6\tiny $\pm0.22$  &  61.9\tiny $\pm0.20$&  55.3\tiny $\pm0.22$ & 	58.3\tiny $\pm0.21$      \\
DSP  &  \textbf{83.2}\tiny$\pm0.14$ &  \textbf{71.0}\tiny$\pm0.16$ & \textbf{76.5}\tiny $\pm0.14$ & \textbf{74.8}\tiny$\pm0.14$  & \textbf{57.9}\tiny$\pm0.19$&  \textbf{70.4}\tiny$\pm0.12$ &  \textbf{62.8}\tiny$\pm0.13$ & \textbf{66.4}\tiny$\pm0.19$\\  \bottomrule
\end{tabular}
\end{table}

\subsection{Evaluation on VL-CMU-CD and PCD Datasets}
Table~\ref{res:vlcmucd} shows the performance of proposed pretraining methods evaluated using DR-TANet on the VL-CMU-CD and PCD dataset. The results show that our proposed pretraining method can surpass the widely-used ImageNet pretraining (Sup-Im) which utilizes millions of images without the use of any additional data. Our method shows 1.3\% and 5.7\% gains on the VL-CMU-CD dataset compared to ImageNet pretraining and randomly initialized weights. Also, when compared to the standard Barlow Twins pretraining, DSP improves the downstream change detection performance by 2\%. This shows the ability of DSP to learn distinctive representations useful for the downstream task of SCD.

We also evaluate the performance of our proposed methods on the PCD dataset where the available unlabeled data is as few as 200 image pairs. Results demonstrate that the proposed pretraining outperforms the standard Barlow Twins pretraining comfortably by a large margin. Our DSP network gains an improvement of $\sim$9\% in the PCD dataset over ImageNet pretraining when compared to 1.3\% gain on the VL-CMU-CD dataset. This reduced performance gains in the VL-CMU-CD dataset can be attributed to the dataset itself where the changes are not so detailed when compared to the PCD dataset. Nevertheless, the large performance gains in these challenging SCD datasets over ImageNet pretraining indicate that our DSP helps to improve the change detection performance even when the changes between the image pairs are predominantly affected by noisy changes caused by illumination, seasonal variations, and view-point differences. Qualitative results of the proposed pretraining method is shown in Figure~\ref{qualitative_results}. Also, precision and recall values for each subset is provided in the Table~\ref{tab:pcd_pr} in the Appendix.

Overall, the evaluation on VL-CMU-CD and PCD datasets shows that our proposed methods on unlabeled data can surpass the Barlow Twins pretraining and the widely used ImageNet pretraining that uses more than a million labeled images. Moreover, it also alleviates the problem of domain shift caused by transfer learning the ImageNet weights pretrained on datasets vastly different from that of SCD datasets. Finally, the proposed invariant prediction objective and change consistency regularization helps to induce temporal invariance during the pretraining stage that minimizes the noise due to significant viewpoint differences.

\begin{table}[t]
  \caption{Ablation study of DSP components on PCD dataset.}
  \label{res:ablation}
  \centering
  \begin{tabular}{lccccc}
    \toprule
    Method & Differencing  & Temporal Consistency & Pretraining  & Label & F1-Score \\
    \midrule
        DSP & \cmark & \cmark & \cmark  & \xmark  & \textbf{66.4}    \\
         & \cmark & {only $\mathcal{L_{CR}}$} & \cmark & \xmark & 65.1\\
         & \cmark & \xmark & \cmark & \xmark   & 64.2 \\
        
         & \xmark & \cmark & \cmark & \xmark & 60.1\\ 
          
         Barlow Twins & \xmark & \xmark & \cmark & \xmark & 58.3\\ \midrule 
        Sup-Im & \xmark & \xmark & \cmark  & 1M & 57.6\\
        Rand Init & \xmark & \xmark & \xmark  & \xmark & 53.5\\
    \bottomrule
  \end{tabular}
\end{table}

\subsection{Ablation Study}
To examine the influence of individual components of our DSP network on downstream SCD performance, we perform the following ablation study. As discussed in Section \ref{methodology}, the two main components of our network are temporal consistency loss $\mathcal{L}_{TC}$ and the feature differencing layer.

Table~\ref{res:ablation} shows the results of the ablation study where we perform all experiments using the PCD dataset.
To analyze the effect of the change consistency regularizer, we perform an experiment by removing the invariant prediction term alone. Results show that there is a drop in performance (1.3\%). Therefore, both the invariant prediction and the change consistency regularizer are important for the DSP model.
The performance drops by 2.2\% when we remove the temporal consistency loss entirely. This confirms the importance of temporal consistency loss for the DSP model as it helps to tackle the presence of noisy changes in the input data by enforcing invariant prediction across intra-views during the pretraining stage.   

Then, to analyze the impact of differencing layer in the self-supervised model, the feature differencing layer in the DSP network is removed and the model is trained on $\mathcal{L}_{SSL}$ alone by maximizing the cross-correlation between the features ($f_0',f_1'$) of the image pairs taken at different times. Training this way is equivalent to the state-of-the-art Barlow Twins method \citep{zbontar2021barlow} applied on SCD. We observe that the Barlow Twins method leads to a $\sim$6\% drop in downstream change detection performance compared to the proposed DSP model which uses differencing layer. This performance drop using Barlow Twins as SSL loss may be due to the diminishing discrimination ability of the model in the downstream SCD task, as maximizing the cross-correlation between the temporal images will make the representations of the image pairs along with that of the changed regions are forced to be closer together. In addition, we found that just adding temporal consistency loss to the Barlow Twins model preserves the change between the temporal images and boosts the change detection performance by 1.8\%. Based on the ablation studies, we conclude that the feature differencing and the temporal consistency loss enables the DSP model to learn discriminatory representations corresponding to the changed regions that are beneficial for SCD.

Finally, results show that pretraining with DSP on in-distribution dataset alone addresses the problem of domain shift and increases the downstream change detection performance by 8.8\% over supervised ImageNet pretraining (Sup-Im) on one million labeled data respectively. Thus, pretraining with DSP resolves domain shift, is label efficient, and leads to increased performance when compared to other pretraining regimes in SCD.  

\begin{table}[t]
    \centering
\caption{Effect of loss balancing parameters on the DSP performance on VL-CMU-CD dataset.} \label{tab:hyp2}
    \begin{tabular}[t]{lllc}
        \toprule
Method & $\alpha$ & $\beta$ & F1-Score   \\ \midrule
DSP    & 100      & 0       & 75.95      \\
      & 500      & 0       & 75.93      \\
      & 1000     & 0       & 75.90       \\ \bottomrule
    \end{tabular}\vline%
    \begin{tabular}[t]{lllc}
        \toprule
Method & $\alpha$ & $\beta$ & F1-Score   \\ \midrule
DSP    & 100      & 1000    & 76.20     \\
      &          & 3000    & 76.50       \\
      &          & 5000    & 76.52     \\ \bottomrule
    \end{tabular}
\end{table}

\par{\bf Hyperparameter Sensitivity Analysis. } 
{Table~\ref{tab:hyp2} reports the sensitivity of the model's performance to change in loss term coefficients (\Eqref{total_loss}).
 To analyze the effect of $\alpha$ on the performance, we vary the $\alpha$ by keeping the $\beta$ constant. We observe that increasing $\alpha$ leads to a limited influence on the final downstream performance. Therefore, we keep the $\alpha$ value fixed to 100 and vary the $\beta$ term to analyze its sensitivity to the performance. Since the scale of the loss is small compared to the other losses $(\mathcal{L}_{IP}, \mathcal{L}_{SSL})$, we increase the value of $\beta$ in thousands to balance these losses in the same scale. We observe that increasing $\beta$ initially boosts the performance by a small margin and later its effect is saturated. We have found that setting $\alpha$ =100 and $\beta$ = 3000 works best (by a small margin) for VL-CMU-CD. Subsequently, we have also obtained excellent results on the PCD dataset using these same values. Therefore, it is evident that the performance of the DSP is less sensitive to the hyperparameters of the loss function.
 Note that we could easily tune these parameters by cross-validation on the validation sets of these two small scene change detection datasets.}


\begin{table}[t]
\centering
\caption{Out-of-distribution performance evaluation (F1-score) of finetuning model using pretraining methods.} \label{res:ood}
\centering
\begin{tabular}{lcc}
\toprule
  Methods  
  &VL-CMU-CD$\to$PCD  &PCD$\to$VL-CMU-CD \\\midrule
  Rand Init & 23.4\tiny $\pm0.52$   & 18.6\tiny $\pm0.32$     \\   
  Sup-Im & 	28.6\tiny $\pm0.21$    & 22.8\tiny $\pm0.12$      \\ 
   Barlow Twins & 	36.6\tiny $\pm0.17$    & \textbf{30.6}\tiny $\pm0.15$    \\
  DSP & \textbf{43.1}\tiny $\pm0.16$    & 27.5\tiny $\pm0.20$   \\
\bottomrule
\end{tabular}
\end{table}

\subsection{Generalization on out-of-distribution data} \label{sec:ood}
In practice, the SCD model has to perform in challenging scenarios where the testing distribution is unknown and drastically different from the one it is trained on. Therefore, the learned representations must generalize well across out-of-distribution data. The PCD dataset is considered OOD data for a model pretrained and finetuned on the VL-CMU-CD dataset and vice versa. Table \ref{res:ood} shows the F1-score of different pretraining methods to out-of-distribution data. Barlow Twins pretrained and finetuned on PCD outperforms DSP whereas the DSP model finetuned on VL-CMU-CD displayed better generalization compared to Barlow Twins method. Results show that the Barlow Twins and DSP model generalizes well to the OOD dataset compared to the model initialized with random weights and ImageNet pretrained weights, indicating that self-supervised pretraining helps learn more generalizable feature representations.

\subsection{Robustness under Natural Corruption} \label{sec:corruptions}
\begin{wrapfigure}[18]{R}{0.5\textwidth}
    \vspace{-18pt}
     \centering
     \includegraphics[width = 0.5\textwidth, trim={0.3cm 0.28cm 1.5cm 0.7cm},clip]{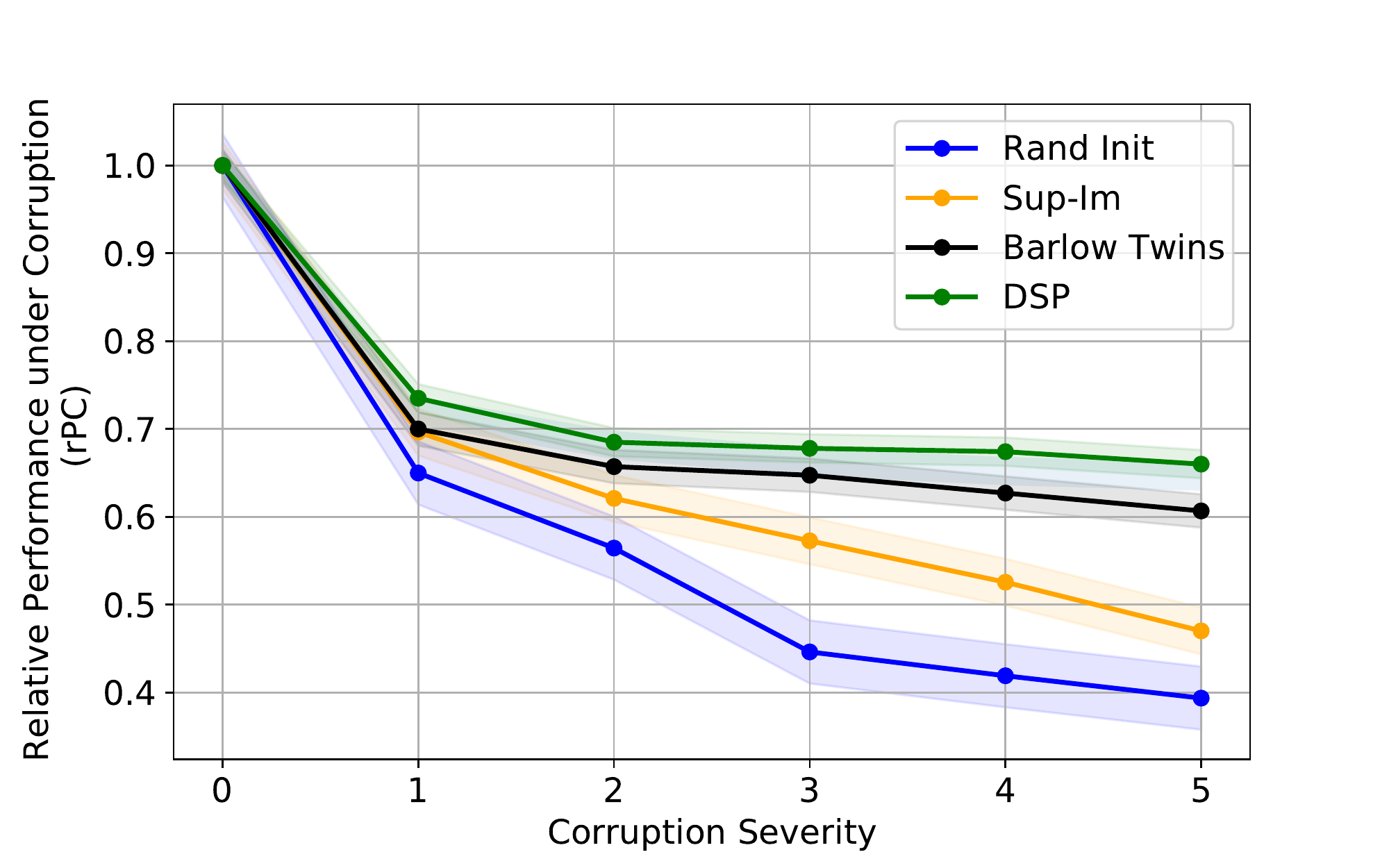}
     \caption{Relative Performance degradation on corrupted images with increasing levels of corruption severity, best viewed on color.}
     \label{fig:Corrup_result}
\end{wrapfigure}
The task of SCD is often applied to the outdoor environment, where the images are subjected to seasonal variations. Therefore, the SCD model must be robust to the common natural corruptions such as illumination, noise, and blur. Here, we evaluate the robustness of SCD models to natural corruptions which has not been addressed in this domain previously. Following \citet{hendrycks2019benchmarking}, we use 15 different natural corruptions applied on the VL-CMU-CD test set to generate VL-CMU-CD-C (examples are shown in Figure~\ref{fig:corruptions} in Appendix).
These corruptions are categorized into 4 categories as noise, blur, weather, and digital. Each corruption category is subjected to five severity levels obtained by varying intensities of corruption. The finetuning model initialized with different pretraining methods is trained using clean VL-CMU-CD while being tested on VL-VMU-CD-C. The metrics mean performance under corruption (mPC) \citep{michaelis2019benchmarking}  and relative performance under corruption (rPC) \citep{michaelis2019benchmarking} are used to evaluate the robustness of models (~\Eqref{eq:corrpmetric} and \Eqref{rcorrp}, respectively). rPC measures the relative degradation of performance on corrupted data with respect to clean data.
\begin{equation}
    \label{eq:corrpmetric}
    mPC = \frac{1}{N_c N_s} \sum _{c=1}^{N_c} \sum _{s=1}^{N_s} P_{c,s}
\end{equation}
where $P_{c,s}$ is the F1-score measure evaluated on VL-CMU-CD-C under c$th$ corruption with severity levels. while $N_c$= 15 and $N_s$= 5 indicate the number of corruptions and severity levels, respectively. 
\begin{equation}
    \label{rcorrp}
     rPC = \frac{mPC}{P_{clean}}
\end{equation}

Figure~\ref{fig:Corrup_result} shows that the drop in performance increases with an increase in the severity of the applied corruption. Moreover, (1) there is a large degradation in performance on the model initialized with supervised Imagenet pretraining (Sup-Im) when subjected to the corrupted test set. (2) Unlike the ImageNet pretrained models that suffer severe performance loss on corrupted images, our proposed pretraining method DSP is more robust to increase in corruption severity as the degradation in performance is gradual. (3) Compared to the standard Barlow Twins, DSP is more robust as it likely learns robust representations that are invariant to noisy changes during model pretraining. Therefore, it is evident that the proposed self-supervised pretraining brings discernible benefits in terms of robustness to natural corruptions when compared to the ImageNet pretraining and the Barlow Twins method.

\begin{table}[tb]
\caption{Performance (F1-score) of pretraining methods evaluated on VL-CMU-CD dataset under varying label availability.} \label{res:limitedlabel}
\setlength\tabcolsep{0pt} 
\centering
\begin{tabular*}{0.7\columnwidth}{@{\extracolsep{\fill}}lcccc} 
\toprule
   Methods  &\multicolumn{4}{c}{Label Fraction}\\
  \cmidrule(r){2-5}
  &1\% & 10\%  & 50\%  & 100\%  \\
\midrule
     Rand Init& 25.2\tiny $\pm0.50$     & 42.3\tiny $\pm0.32$      & 54.5\tiny $\pm0.72$     & 70.8\tiny $\pm0.51$  \\
   Sup-Im&29.5\tiny $\pm0.20$    & 41.1\tiny $\pm0.18$     & 60.1\tiny $\pm0.18$    & 75.2\tiny $\pm0.15$ \\

   Barlow Twins & 56.9\tiny $\pm0.14$    &61.7\tiny $\pm0.16$     & 68.5\tiny $\pm0.15$ & 74.5\tiny $\pm0.12$ \\
  DSP & \textbf{58.7}\tiny $\pm0.30$    & \textbf{65.9}\tiny $\pm0.25$    & \textbf{71.8}\tiny $\pm0.14$    & \textbf{76.6}\tiny $\pm0.14$ \\
\bottomrule
\end{tabular*}
\end{table}

\subsection{Efficiency under Limited Labels} \label{sec:limitedlabels}
The availability of large annotated data remains a critical challenge in SCD due to the high cost of acquiring manual annotations. Therefore, the SCD model needs to demonstrate steady performance when the availability of labeled data is limited. Table \ref{res:limitedlabel} shows the performance of different pretraining under limited labels setting. Different percentages of labeled data (1\%, 10\%, 50\%, and 100\%) are sampled in a class-balanced manner from the training split of VL-CMU-CD. The finetuning performance of the DSP pretraining method with varying quantities of labeled data is evaluated using the VL-CMU-CD test set. Our proposed pretraining method (DSP) outperforms the widely used Imagenet pretraining by a large margin across all limited label scenarios. The performance drop of Sup-Im is more significant when the amount of labeled data is 10\% or less compared to our method, and then the gap decreases as the availability of labeled data increases. When compared to the self-supervised Barlow Twins pretraining, our method (DSP) increases the performance of the change detection model to a greater extent when the availability of the labeled data is scarce due to learning more generic feature representations using differencing based self-supervised learning.

In general, we observe self-supervised pretraining enhances the downstream performance in SCD. Overall, compared to ImageNet and Barlow Twins pretraining, the proposed DSP method generally increases the performance and the robustness and generalization of the SCD model to a larger extent in many real world scenarios where the images are affected by challenging conditions. This can be attributed to the learned representations of the changed features through differencing, the proposed invariant prediction, and the change consistency regularization which all together help DSP to learn more generalizable features that further contribute to the increase in robustness of an SCD model.

\section{Conclusion and Future work} \label{sec:conclusion}
We proposed a novel \textit{differencing based self-supervised pretraining method (DSP)} for scene change detection that learns temporally consistent features inherent to the data in a self-supervised manner. Furthermore, we find that explicitly incorporating temporal invariance into the self-supervised pretraining resulted in a more tolerant strategy to punish the irrelevant changes caused due to large viewpoint differences. With extensive experiments on two challenging SCD datasets, we demonstrated the superiority of the DSP over the self-supervised Barlow Twins and the widely used ImageNet pretraining without any additional data. Our results also demonstrate the robustness of DSP to natural corruptions, out-of-distribution generalization, and its efficiency under limited annotations. Therefore, we believe that our findings in this work can be harnessed to increase the performance and robustness of SCD where obtaining the labeled data is scarce and expensive. Although our approach reduces the dependency of the SCD models to large-scale labeled data, one possible limitation is that the task of SCD is not entirely unsupervised. In the future, we intend to extend the proposed self-supervised approach to tackle the problem of unsupervised change detection. 


\bibliography{collas2022_conference}
\bibliographystyle{collas2022_conference}

\appendix
\section{Appendix}
\subsection{Self-supervised pre-training setup} \label{setup}

\textbf{Dataset pre-processing.} 
For the PCD dataset, original images are cropped into
224×224. By sliding 56 pixels in width and data augmentation of plane rotation, each image pair is expanded into 60 patches with a 224×224 resolution. In total, 12000 image pairs are generated. As the input, the image pairs will be resized into 256×256.
For the VL-CMU-CD dataset, we follow the random training and testing splits reported in \cite{guo2018learning, chen2021dr}. Nine hundred thirty-three image pairs (98 sequences) for training and 429 (54 sequences) for testing are resized into a 256×256 resolution. Note that only images belonging to the train set (without labels) are used to train the DSP model. For data pre-processing, training, and validation split, we follow the steps mentioned in \cite{chen2021dr} for pretraining in VL-CMU-CD and PCD datasets.  

\textbf{Architecture.} The proposed DSP method has a Siamese architecture that consists of ResNet50 \cite{he2016deep} (without the final classification layer) as a feature extractor followed by a projector network. We use dilated convolutions and reduce the input image size by a stride of 16 to output feature vectors of size $16 \times 16 \times 2048$. Then, the feature vectors are passed to the projection head by applying a 2-D adaptive average pooling. The projector network has two linear layers, each with a hidden layer size of 512 output units. Owing to the high computational requirements, the output of the projector network was modified to generate embeddings of size $1 \times 256$. The first layer of the projector is followed by a batch normalization layer and rectified linear units. Then, an absolute difference is applied to the output embedding to get difference embeddings. The Barlow Twins loss function is applied on the difference embeddings of size $1 \times 256$ to generate a cross-correlation matrix of shape $256 \times 256$. Additionally, as discussed in Section~\ref{methodology}, we incorporate Invariant prediction and Change Consistency regularization loss in our DSP framework. As shown in Figure~\ref{fig:method}, the output of the projector before the difference is used to calculate the invariant prediction loss while the change consistency regularizer is applied to the output of the feature extractor.    

\textbf{Data Augmentation.} Our image augmentation pipeline consists of the following transformations: Image resizing to 256 × 256, color jittering, converting to grayscale, and Gaussian blurring. Except for resizing, the other transformations are applied randomly, with some probability. The random crop is not considered when pre-processing the change detection datasets as the presence of changed regions between an image pair taken at different times is much smaller and random compared to the unchanged regions.

\textbf{Training and Optimization.} We follow the optimization protocol described in Barlow Twins. We use the LARS optimizer \citep{you2017large} and train for 400 epochs with a batch size of 16 on two NVIDIA RTX-2080 Ti GPU. We use a learning rate of 0.003, multiply the learning rate by the batch size, and divide it by 256. The learning rate is reduced by a factor of 1000 using a cosine decay schedule \citep{loshchilov2016sgdr}. We use a weight decay parameter of 1x10$^-6$. Finally, the loss balancing parameters $\alpha$ and $\beta$ used for training our pretraining model is 100 and 3000 respectively. We tuned these parameters by cross-validation on the validation sets of the VL-CMU-CD dataset and balanced these losses to the same scale.
For self-supervised Barlow Twins training, we followed the exact data preprocessing, training and optimization procedure.

\subsection{Change Detection setup}\label{changedetection_setup}
We evaluate the proposed self-supervised pretraining methods by finetuning them to a downstream task of SCD. DR-TANet \citep{chen2021dr}, a state-of-the-art SCD network is considered for finetuning. DR-TANet \citep{chen2021dr} is selected because it achieves state-of-the-art results on SCD datasets. It employs an encoder-decoder architecture that incorporates a temporal attention module to exploit the similarity and dependency of feature maps at two temporal channels. We considered ResNet50 as an encoder for pretraining our DSP model. Therefore, to keep the consistency throughout the experiments, we used ResNet50 \citep{he2016deep} as a feature extractor for finetuning the pre-trained model on both of these networks. During finetuning, the data pre-processing, training, and testing protocols followed by DR-TANet were replicated. We considered a batch size of 8 while training the DR-TANet on VL-CMU-CD and PCD datasets. While training on PCD, we reduced the dependency-scope size of the DR-TANet to 1x1 and trained the model in the lowest setting owing to limited GPU memory and longer training time. The upsampling layer is implemented by bilinear interpolation and the classifier layer is a convolutional layer with kernel size 1. The optimizer Adam is adopted. The learning rate is set to 0.001 with $\beta_1$ = 0.9, $\beta_2$ = 0.999.  We train the finetuning model for 150 epochs on VL-CMU-CD and 100 epochs on the PCD dataset till convergence. For evaluation, we followed the same procedure as followed by the DR-TANet. Similar to DR-TANet, a threshold of 0.5 is used for the calculation of true positive, false positive, true negative, and false negative.
\begin{wrapfigure}[20]{R}{0.6\textwidth}
    \vspace{-18pt}
     \centering
     \includegraphics[width = 0.58\textwidth, trim={0.3cm 0.28cm 0.7cm 0.7cm},clip]{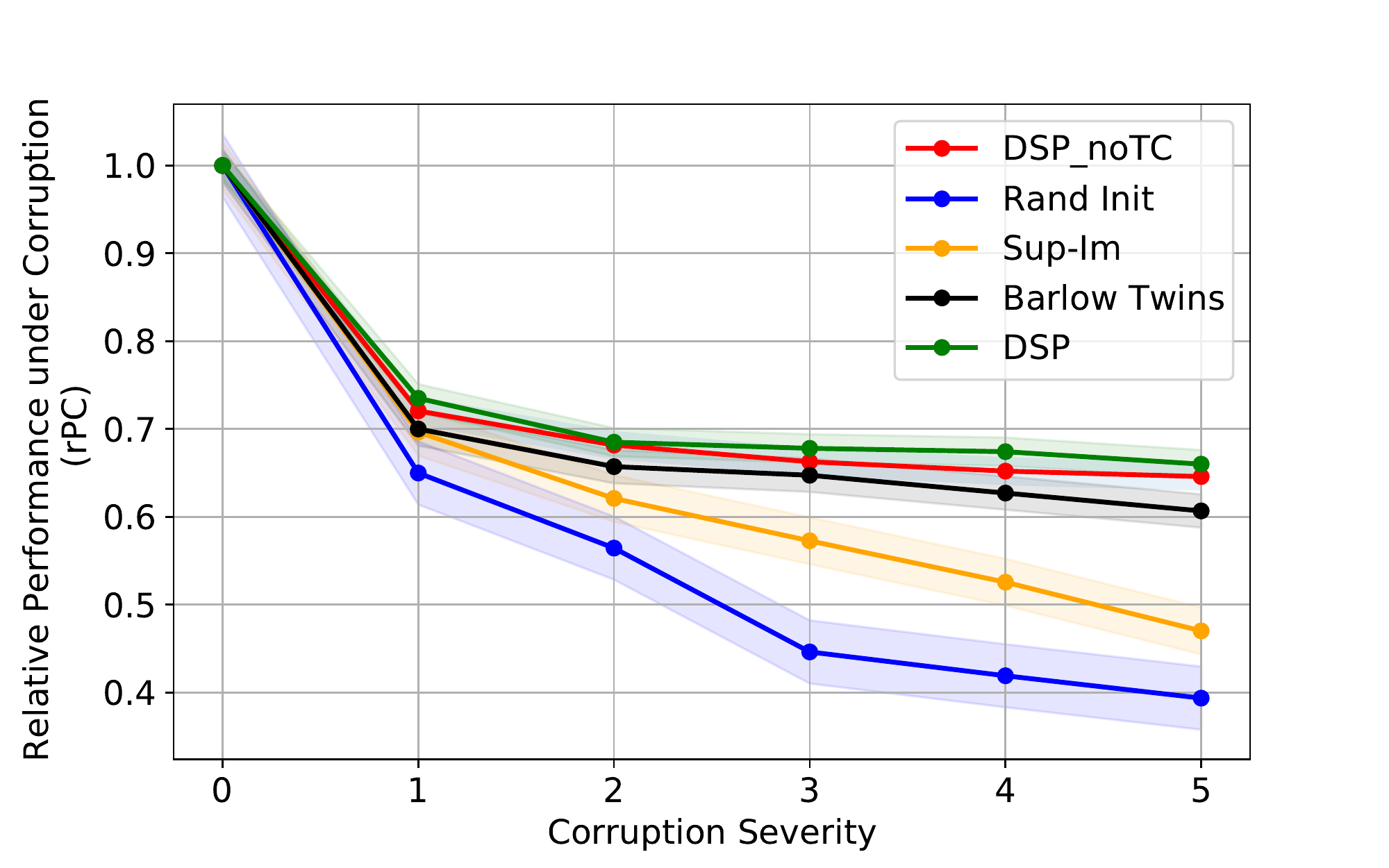}
     \caption{Relative Performance degradation on corrupted images with increasing levels of corruption severity, best viewed on color.}
     \label{fig:Corrup_result_appendix}
\end{wrapfigure}

\subsection{Evaluation under Natural Corruptions}
In Figure~\ref{fig:Corrup_result_appendix}, we show results similar to the performance of different pretraining methods on corrupted images with increasing levels of corruption severity but
with an additional set of experiments shown in Section~\ref{res:ablation}. When compared to the Barlow Twins \citep{zbontar2021barlow}, ImageNet pretraining, and random initialization, our proposed method DSP is more robust and less susceptible to corrupted images. Moreover, to analyze the impact of the temporal consistency loss, we evaluated the DSP model under corruption without the temporal consistency loss. The result shows that the robustness of the DSP is slightly affected without the temporal consistency loss compared to the DSP with temporal consistency. Thus the combination of invariant prediction and change consistency regularizer not only boosts the performance of the downstream scene change detection but also brings robustness under natural corruption scenarios.

\subsection{Qualitative Evaluation} 
\begin{figure*}[t]
\centering
\stackunder[5pt]{
\includegraphics[width=.35\textwidth,height=1.8cm]{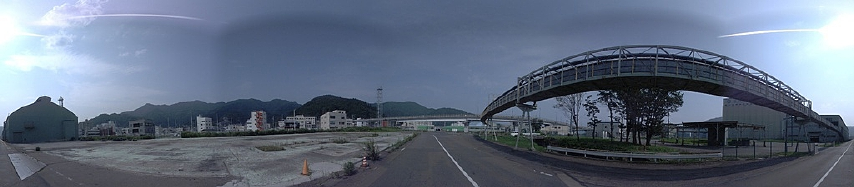}}{Image from Tsunami subset T0}\hspace{0.1cm}%
\stackunder[5pt]{
\includegraphics[width=.35\textwidth,height=1.8cm]{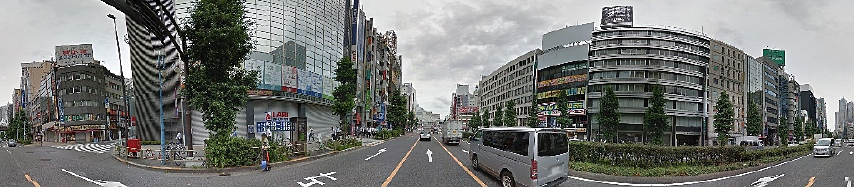}}{Image from GSV subset T0}\hspace{0.1cm}
\\
    [\smallskipamount]
    \stackunder[5pt]{
\includegraphics[width=.35\textwidth,height=1.8cm]{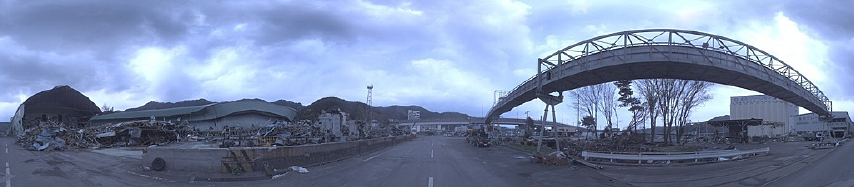}}{Image from Tsunami subset T1}\hspace{0.1cm}%
\stackunder[5pt]{
\includegraphics[width=.35\textwidth,height=1.8cm]{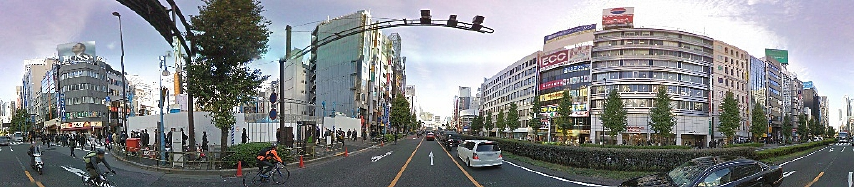}}{Image from GSV subset T1}\hspace{0.1cm}
\\
    [\smallskipamount]
    \stackunder[5pt]{
\includegraphics[width=.35\textwidth,height=1.8cm]{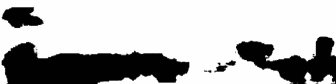}}{Rand Init}\hspace{0.1cm}%
\stackunder[5pt]{
\includegraphics[width=.35\textwidth,height=1.8cm]{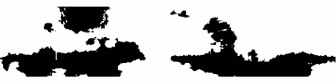}}{Rand Init}\hspace{0.1cm}
\\  [\smallskipamount]
    \stackunder[5pt]{
\includegraphics[width=.35\textwidth,height=1.8cm]{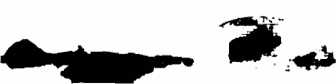}}{Sup-Im}\hspace{0.1cm}%
\stackunder[5pt]{
\includegraphics[width=.35\textwidth,height=1.8cm]{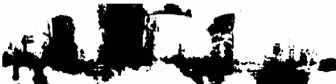}}{Sup-Im }\hspace{0.1cm}
\\  [\smallskipamount]
   \stackunder[5pt]{
\includegraphics[width=.35\textwidth,height=1.8cm]{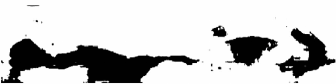}}{Barlow Twins}\hspace{0.1cm}%
\stackunder[5pt]{
\includegraphics[width=.35\textwidth,height=1.8cm]{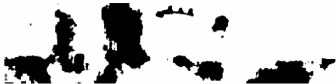}}{Barlow Twins}\hspace{0.1cm}
\\  [\smallskipamount]
   \stackunder[5pt]{
\includegraphics[width=.35\textwidth,height=1.8cm]{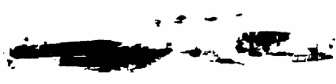}}{DSP}\hspace{0.1cm}%
\stackunder[5pt]{
\includegraphics[width=.35\textwidth,height=1.8cm]{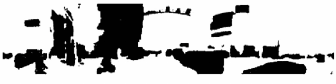}}{DSP }\hspace{0.1cm}
\\  [\smallskipamount]
   \stackunder[5pt]{
\includegraphics[width=.35\textwidth,height=1.8cm]{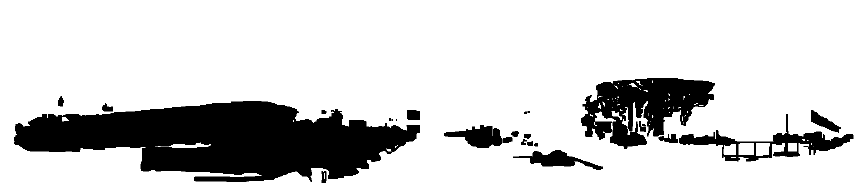}}{Ground truth}\hspace{0.1cm}%
\stackunder[5pt]{
\includegraphics[width=.35\textwidth,height=1.8cm]{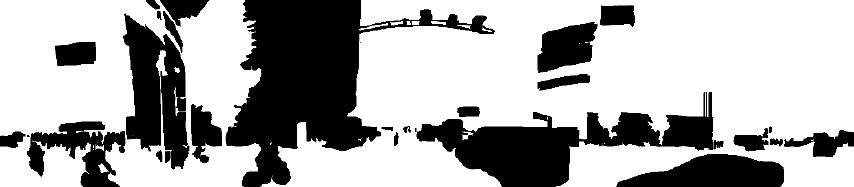}}{Ground truth}\hspace{0.1cm}
\caption{The qualitative results of the pretraining methods on PCD dataset. Results on Tsunami subset (left) and GSV subset (right);}\label{qualitative_results}
\end{figure*}

In Figure~\ref{qualitative_results}, we compare the qualitative results produced by different pretraining methods on the PCD dataset. It can be seen that our proposed DSP pretraining can capture more detailed and small changes, such as tree branches, and pedestrians better when compared to other pretraining methods. This is due to the ability of the DSP to learn discriminate representations that are useful for the downstream task of SCD.\\

\begin{table}[!ht]
\caption{Performance of DR-TANet model trained and tested on PCD dataset using different pretraining methods.}
  \label{tab:pcd_pr}
  \centering
  \resizebox{\textwidth}{!}{
\begin{tabular}{lccccccccc}
 \toprule
\multirow{3}{*}{Methods} & \multicolumn{9}{c}{PCD dataset} \\  \cmidrule(r){2-10}
 &  \multicolumn{3}{c}{Tsunami} & \multicolumn{3}{c}{GSV} & \multicolumn{3}{c}{Average} \\ \cmidrule(r){2-10} 
 & Precision & Recall & F1-score & Precision & Recall & F1-score & Precision & Recall & F1-score  \\ \midrule
Rand Init &  65.7\tiny$\pm0.26$ & 61.9\tiny$\pm0.24$  & 63.4\tiny$\pm0.25$ &42.3\tiny$\pm0.26$ &39.5\tiny$\pm0.25$ &40.7\tiny$\pm0.21$  &  54.0\tiny$\pm0.27$ &  50.7\tiny$\pm0.26$ &53.5\tiny$\pm0.24$  \\
Sup-Im  &  70.2\tiny$\pm0.13$ & 67.5\tiny$\pm0.12$  & 68.7\tiny$\pm0.13$ &  53.0\tiny$\pm0.11$ &41.7\tiny$\pm0.13$ & 46.5\tiny$\pm0.12$ & 61.6\tiny$\pm0.13$  &  54.6\tiny$\pm0.15$ & 57.6\tiny$\pm0.12$ \\
Barlow Twins & 75.2\tiny$\pm0.22$ &67.6\tiny$\pm0.25$ &70.9\tiny $\pm0.18$   &48.6\tiny $\pm0.18$   &42.9\tiny $\pm0.20$    & 45.6\tiny $\pm0.22$  &  61.9\tiny $\pm0.20$&  55.3\tiny $\pm0.22$ & 	58.3\tiny $\pm0.21$      \\
DSP & \textbf{77.0}\tiny$\pm0.10$  & \textbf{72.7}\tiny$\pm0.15$ & \textbf{74.8}\tiny$\pm0.14$ &  \textbf{63.8}\tiny$\pm0.14$ & \textbf{53.0}\tiny$\pm0.13$ & \textbf{57.9}\tiny$\pm0.19$&  \textbf{70.4}\tiny$\pm0.12$ &  \textbf{62.8}\tiny$\pm0.13$ & \textbf{66.4}\tiny$\pm0.19$\\  \bottomrule
\end{tabular}}
\end{table}

\begin{figure*}[t]
\centering
\stackunder[5pt]{
\includegraphics[width=.18\textwidth,height=1.8cm]{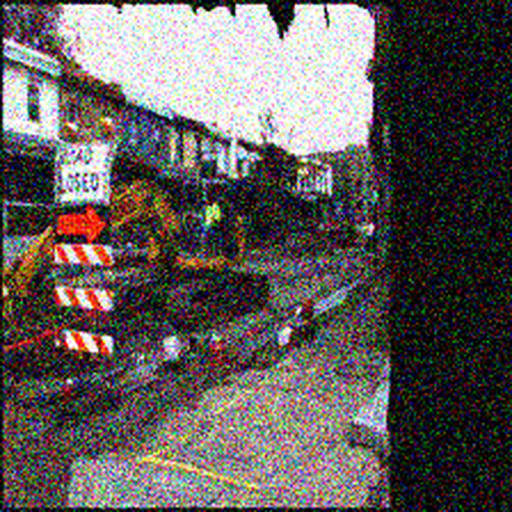}}{Gaussian Noise}\hspace{0.1cm}%
\stackunder[5pt]{
\includegraphics[width=.18\textwidth,height=1.8cm]{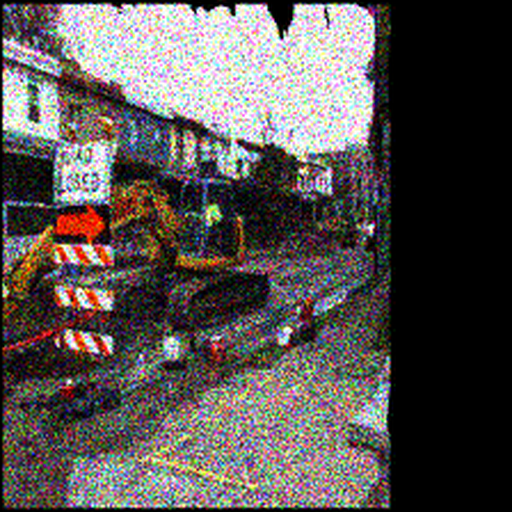}}{Shot Noise}\hspace{0.1cm}%
\stackunder[5pt]{
\includegraphics[width=.18\textwidth,height=1.8cm]{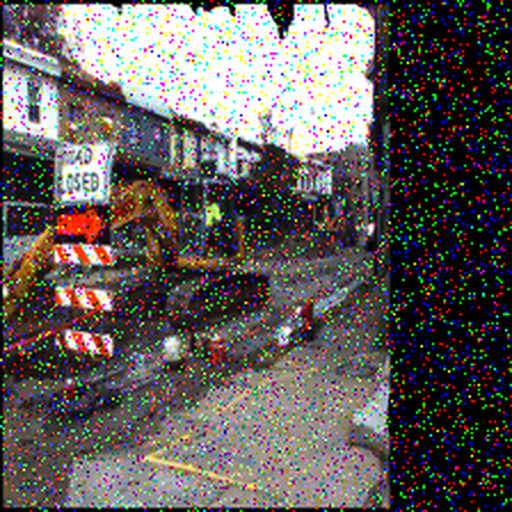}}{Impulse Noise}\hspace{0.1cm}%
\stackunder[5pt]{
\includegraphics[width=.18\textwidth,height=1.8cm]{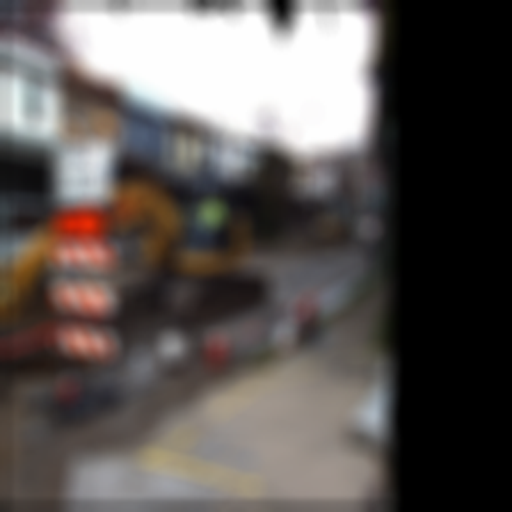}}{Defocus Blur}\hspace{0.1cm}%
\stackunder[5pt]{
\includegraphics[width=.18\textwidth,height=1.8cm]{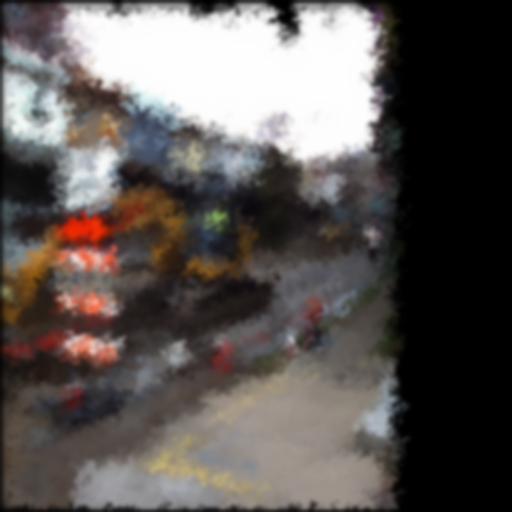}}{Glass Blur}\\
    [\smallskipamount]
\stackunder[5pt]{    \includegraphics[width=.18\textwidth,height=1.8cm]{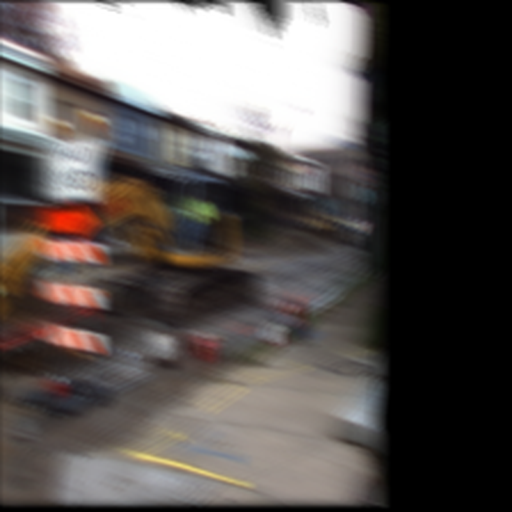}}{Motion Blur}\hspace{0.1cm}%
\stackunder[5pt]{    \includegraphics[width=.18\textwidth,height=1.8cm]{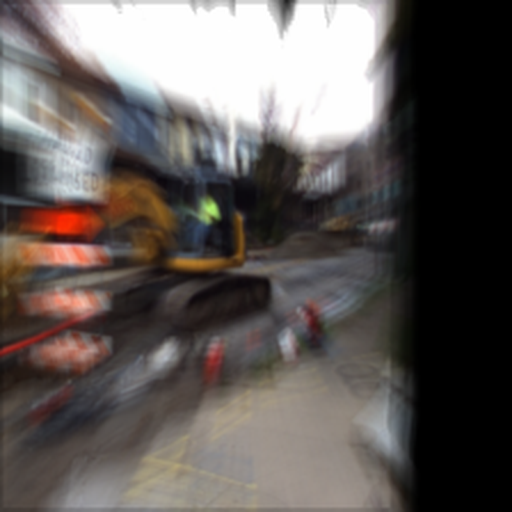}}{Zoom Blur}\hspace{0.1cm}%
\stackunder[5pt]{    \includegraphics[width=.18\textwidth,height=1.8cm]{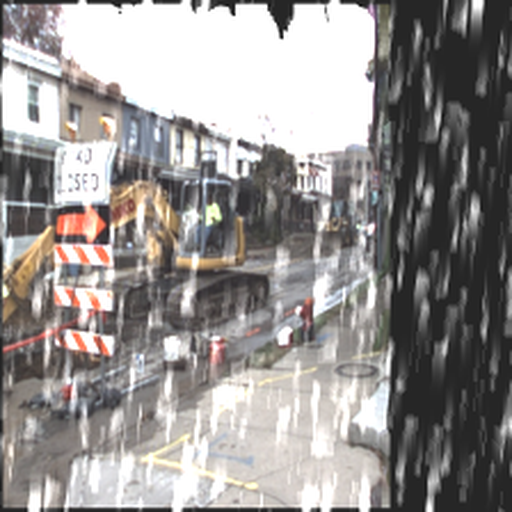}}{Snow}\hspace{0.1cm}%
\stackunder[5pt]{    \includegraphics[width=.18\textwidth,height=1.8cm]{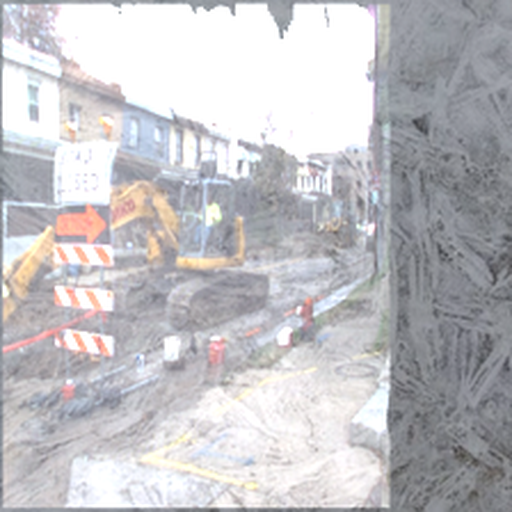}}{Frost}\hspace{0.1cm}%
\stackunder[5pt]{    \includegraphics[width=.18\textwidth,height=1.8cm]{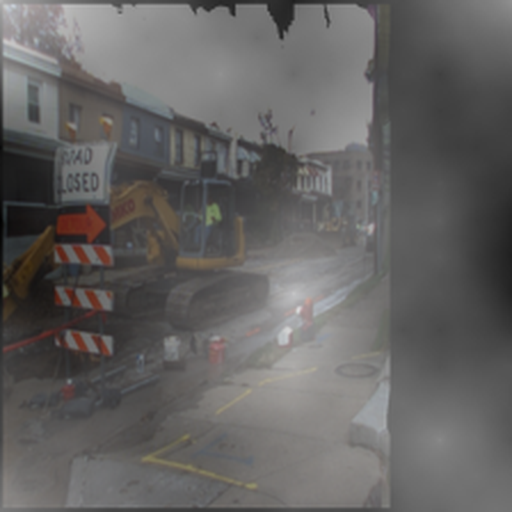}}{Fog}\\
    [\smallskipamount]
\stackunder[5pt]{    \includegraphics[width=.18\textwidth,height=1.8cm]{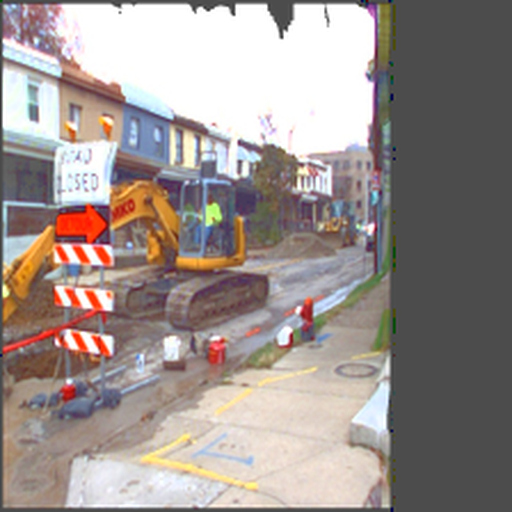}}{Brightness}\hspace{0.1cm}%
 \stackunder[5pt]{   \includegraphics[width=.18\textwidth,height=1.8cm]{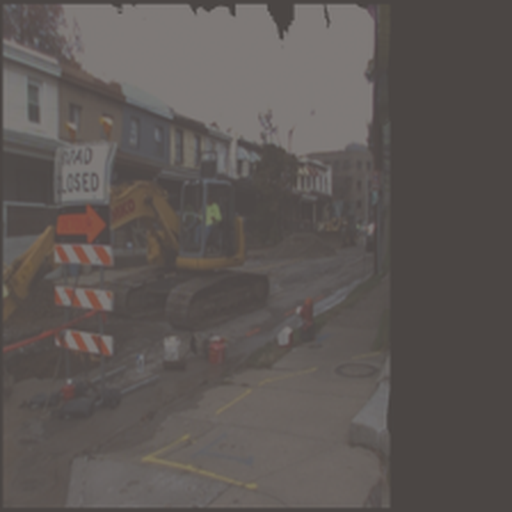}}{Contrast}\hspace{0.1cm}%
 \stackunder[5pt]{   \includegraphics[width=.18\textwidth,height=1.8cm]{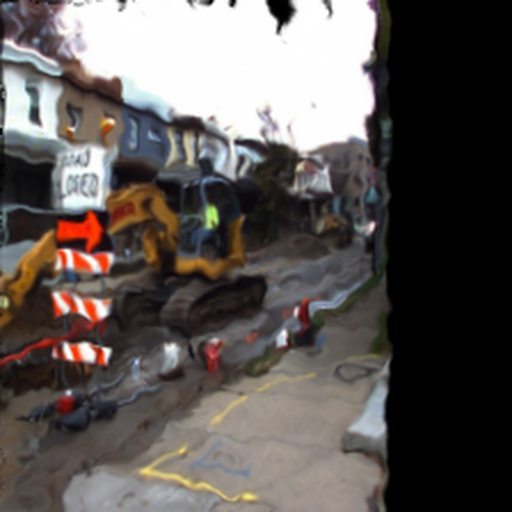}}{Elastic Transform}\hspace{0.1cm}%
\stackunder[5pt]{    \includegraphics[width=.18\textwidth,height=1.8cm]{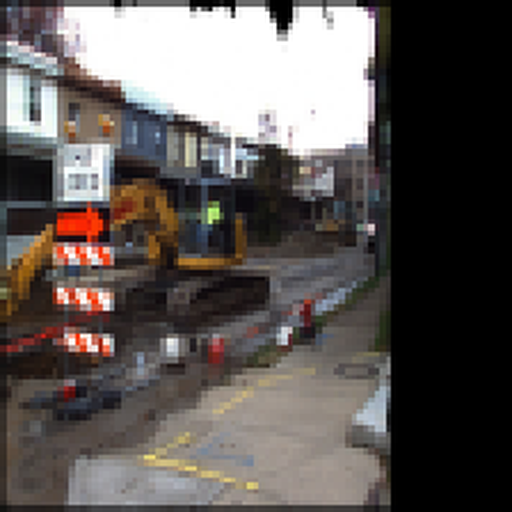}}{Pixelate}\hspace{0.1cm}%
\stackunder[5pt]{    \includegraphics[width=.18\textwidth,height=1.8cm]{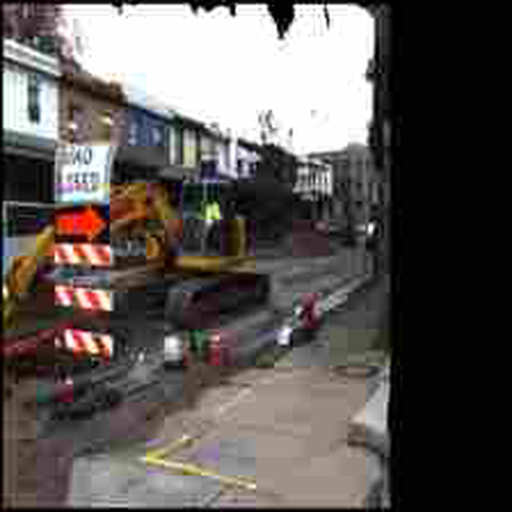}}{JPEG Compression}
\caption{VL-CMU-CD test set is exposed to 15 types of artificially generated natural corruptions \citep{hendrycks2019benchmarking} with five levels of corruption severity. This figure shows a randomly selected image from the VL-CMU-CD test set with severity 3—best viewed on color.}\label{fig:corruptions}
\end{figure*}

\subsection{Relation to Traditional scene change detection methods}
Traditional change detection methods such as image subtraction \citep{hussain2013change}, and Change Vector Analysis (CVA) \citep{malila1980change} recognize scene changes by calculating the pixel-wise algebraic difference between the pair of temporal images.  However, most of these traditional change detection methods work on hand-crafted features their change detection performance is very low and they are highly susceptible to pixel level changes. Recently, Neural Network-based SCD methods perform quite well in challenging scenarios because of their ability to extract multi-scale features and consistently outperform the performance of the traditional change detection methods. In our method, the self-supervised strategy is driven by differentce representations and temporal consistency is used to learn the visual feature representations, which are more consistent and discriminative to directly compare the difference. Thus, we hypothesize that our DSP pretraining leads to a better Difference Image where changed areas are significantly enhanced and unchanged ones are suppressed. This also alleviates the difficulty of the difference image generation for the final change map. Therefore, our DSP method can be used in combination with other traditional SCD methods such as CVA or image differencing to generate the change map in an unsupervised manner.

\end{document}